# An Improved CNN-based Neural Network Model for Fruit Sugar Level Detection

Boyang Deng*†, Xin Wen‡, Zhan Gao‡

†Department of Biosystems & Agricultural Engineering, Michigan State University, East Lansing, MI 48824, United States.

‡School of Science, Beijing Jiaotong University, Beijing 100044, China

**ABSTRACT:** Artificial Intelligence (AI) is widely used in image classification, recognition, text understanding, and natural language processing, leading to significant advancements. In this paper, we introduce AI into the field of fruit quality detection. We designed a regression model for fruit sugar level estimation, utilizing an Artificial Neural Network (ANN) based on the visible/near-infrared (V/NIR) spectra of fruits. After analyzing the fruit spectra, we proposed an innovative neural network structure: the lower layers consist of a Multilayer Perceptron (MLP), a middle layer features a 2-dimensional correlation matrix, and the upper layers contain several Convolutional Neural Network (CNN) layers. Using fruit sugar levels as the detection target, we collected data from two fruit types, Gan Nan Navel and Tian Shan Pear, and conducted separate experiments to compare their results. To assess the reliability of our dataset, we first applied Analysis of Variance (ANOVA). We then explored various strategies for processing spectral data and evaluated their impact. Additionally, we employed Wavelet Decomposition (WD) for dimensionality reduction and a Genetic Algorithm (GA) to identify optimal features. We compared the performance of Neural Network models with traditional Partial Least Squares (PLS) models, and specifically evaluated our proposed MLP-CNN structure against other traditional neural network architectures. Finally, we introduced a novel evaluation metric based on the dataset's standard deviation (STD) to assess detection performance, demonstrating the feasibility of using an artificial neural network model for nondestructive fruit sugar level detection.

## 1. Introduction

Since the last century, researchers have attempted to apply V/NIR spectroscopy technology to the field of fruit quality detection. However, traditional experimental systems were large, with inconvenient parameter adjustments, costly and slow devices, and a lack of stability. In recent years, rapid advancements in AI and deep learning have extended the concept of artificial neural networks, demonstrating that deeper networks with more layers often outperform shallower networks. Artificial neural networks excel in learning abstract features and offer robust self-feedback and adjustment capabilities. In this paper, we leverage artificial neural networks, incorporating the principles of deep learning, allowing for flexibility in the number of network layers in our regression model. We developed an innovative V/NIR spectroscopy-based approach for detecting fruit sugar levels, constructing a regression model tailored to this purpose.

## 2. Relative Works

In the field of fruit quality detection, the United States, Japan, and Europe have been dedicated to non-destructive fruit detection methods since the last century. McGlone er al. (1998) used NIR spectroscopy (800-1100 nm) to non-destructively assess kiwi fruit maturity, developing a multivariate model focused on dry matter content and sugar levels. Evaluation metrics included the coefficient of determination ($R^2$) and root mean square error of prediction (RMSEP), with results showing $R^2$ = 0.90 and RMSEP = 0.42 for dry matter content, and $R^2$ = 0.90 and RMSEP = 0.39 for sugar levels. We propose an innovative evaluation standard beyond these metrics, which will be discussed later.

Kim et al. (2000) applied visible/near-infrared (V/NIR) spectroscopy to assess kiwi fruit maturity, examining the effects of growing environment and storage time on maturity. Linear and nonlinear models were developed, with the nonlinear models showing superior performance. Our model, a nonlinear neural network, produced experimental results consistent with these findings.

In another study, McGlone et al. (2002) compared density-based methods and V/NIR approaches for detecting dry matter and sugar levels in kiwi fruit, finding comparable performance between the two. We chose the V/NIR approach for fruit sugar level detection. Els et al. (2010) investigated how differences in apple samples—such as production location and exposure time—affected the accuracy of sugar detection models, identifying key wavelength distinctions at 970 nm, 1170 nm, and 1450 nm. Their findings indicated that sample diversity improves model stability. For our study, we collected 300 samples each of navel oranges and pears, the maximum within our capacity.

In deep learning, Rosenblatt et al. (1957) introduced the concept of the perceptron, which enables binary classification of multidimensional data and can learn and update weights using the gradient descent algorithm, a technique foundational to our work. Minsky et al. (1969) later showed that the perceptron, as a linear model, could only solve linear classification problems and was insufficient for solving complex tasks like the XOR problem. Hinton et al. (1986) developed the backpropagation algorithm, enabling multilayer perceptrons (MLPs) to handle nonlinear classification by using the sigmoid function for nonlinear mapping. This inspired the MLP component in our model.

Yann LeCun et al. (1998) proposed the convolutional neural network model LeNet, which achieved high accuracy in Arabic numeral recognition. AlexNet, introduced by Alex Krizhevsky et al. (2012), was a more advanced CNN model, winning the ILSVRC-2012 competition with a top 5 test error rate of 15.3%, compared to 26.2% for the second-best entry. Our CNN architecture draws inspiration from AlexNet. Innovations in neural networks continued with models like GoogleNet, which influenced our use of small convolutional kernels, and Kaiming He et al. (2015) proposed DeepResidualNet, a 150-layer CNN model that demonstrated how increased network depth could enhance recognition performance when well-designed. Accordingly, our neural network model includes 12 layers as a laboratory prototype.

Recent deep learning research has primarily focused on image classification, recognition, and natural language processing. We were inspired by Can Wang et al. (2018), which applied neural networks and a V/NIR approach to detect soil moisture content. This study transformed 1-dimensional spectral data into a 2-dimensional information matrix, guiding our approach. However, there remains a need for specialized deep learning models tailored to effectively learn fruit spectral features.

## 3. Experiment Objects and Operating Methods

### 3.1 Experiment Objects

In this paper, we selected the Gan Nan Navel Orange and Tian Shan Pear as research subjects. Using Gan Nan Navel Orange as an example, we chose 300 samples, each weighing between 200g and 300g, with normal shape, uniform size, and no visible surface scars. First, each fruit sample was cleaned individually with a damp towel, labeled from 1 to 300. The samples were then stored in an isothermal environment at 24°C for 24 hours.

### 3.2 Spectrum Pick Methods

In this paper, we primarily use experimental apparatus such as the USB2000+ Micro Commercial Light Ray Spectroscope (Ocean Optics Inc., USA), a 50W halogen lamp, and a standard diffuse reflection whiteboard. First, the 50W halogen lamp light source undergoes a 30-minute preheating. Then, we set parameters in the spectral acquisition software as follows: integration time is 100ms, averaging time is 32, and the smoothing factor is 4. Each Navel Orange sample is measured at four evenly spaced points along the fruit equator. The experimental setup is shown in Figure 1.

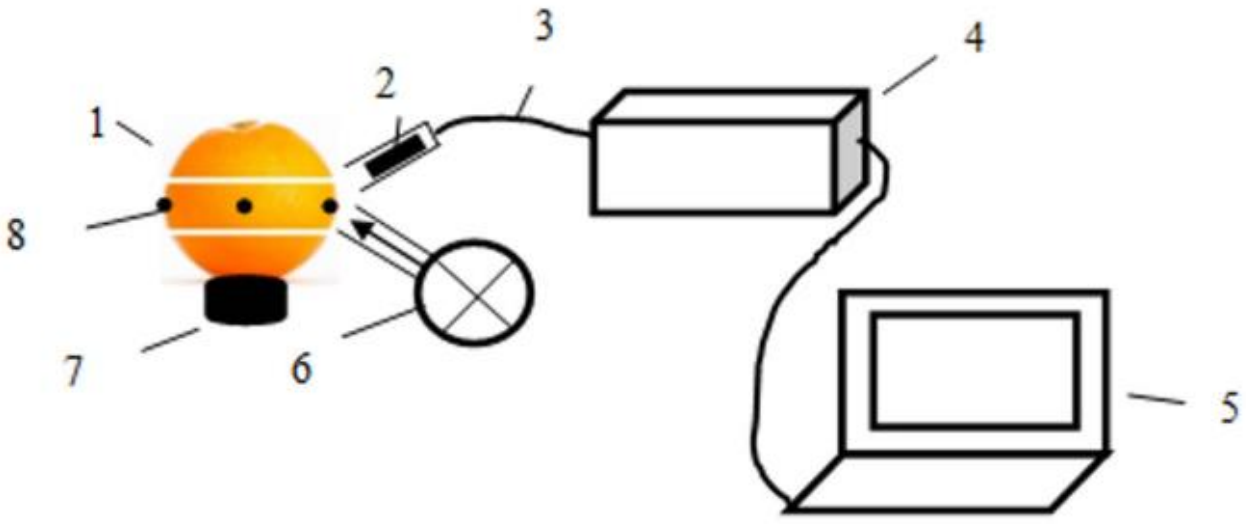

1. Navel orange; 2. Fiber optic probe; 3. Optical fiber; 4. Spectrometer; 5. PC 6. Light source; 7. Supporter; 8. Fruit equatorial sampling point.

Figure 1. Experimental setup for V/NIR spectroscopy.

### 3.3 Real Sugar Value Detection Method

Sugar values for the Navel Orange samples were measured using the LB32Y handheld spectroscope. Pulp juice was extracted from four sample points using a pipette and placed on the sample board. The scale position was recorded, and the spiral was adjusted until distinct blue and white bands appeared in the spectroscope view. The average of these measurements was taken as the true sugar value for each fruit sample.

### 3.4 Data Process and Model Evaluation

In this paper, we used the chemometrics software Unscrambler X10.4 (CAMO, Trondheim, Norway) and TensorFlow 1.8 for data processing and model building. Model performance was validated using cross-validated root mean square error (RMSECV), model prediction coefficient of determination ($R^2$), and a new evaluation standard we proposed. A lower RMSECV or higher $R^2$ indicates stronger predictive ability of the model. We will discuss the new evaluation standard shortly. Unless otherwise specified, RMSECV is used to represent the test set (or validation set; in our case, the validation set is equivalent to the test set). Additionally, as cross-validation results serve as our standard, standard deviation (STD) refers to the overall dataset sugar value STD by default.

## 4. Data Analysis and Model Constructing

Deep learning is commonly used to solve problems such as image labeling, text analysis, and natural language recognition. Compared to these, our research on fruit sugar level detection presents unique challenges, as humans can intuitively understand the meaning of images, text, or language, but not fruit spectra. For example, in dog recognition, we can easily identify dogs within an image dataset and even distinguish between dog breeds after minimal training. With text, we understand the content and can design deep learning models informed by our insights, enabling supervised learning of text features. Similarly, human understanding of spoken language aids in the development of conversational robots.

In contrast, fruit spectra data relies on the Lambert-Beer law to construct a mathematical model where the independent variable is fruit sugar content and the dependent variable is absorbance, linked to hydrogen group interactions with light. For our dataset, spectra from similar fruit types show consistent features, including comparable numbers of peaks and troughs. However, high-sugar and low-sugar fruits may both exhibit prominent peaks, and high-sugar fruits may have either steep or gentle peaks in the same spectral segments. Unlike images or text, we cannot intuitively discern the relationship between spectra and sugar levels, nor do we know which spectral wavelengths specifically represent hydrogen group absorption, as energy transitions occur across multiple segments.

External factors, such as experimental conditions, equipment, and data collection methods, also influence the observed spectra, making them differ from theoretical predictions. Consequently, we include a data preprocessing stage before training our neural networks, which helps to filter noise and identify relevant spectral ranges. While deep learning can autonomously learn data features, it may also capture noise, which can interfere with model training and reduce predictive accuracy.

We applied 10-fold cross-validation for traditional models and 5-fold cross-validation for neural network models. In total, we have 300 samples each of Navel oranges and pears. For 5-fold cross-validation, the validation set size is 60 samples, with the remaining samples used as the training set, giving a training-to-total sample ratio of 0.8 per fruit. For 10-fold cross-validation, the validation set size is 30 samples, resulting in a training-to-total sample ratio of 0.9 per fruit.

For example, we built an initial PLS model using 1,600 wavelength points per pear, where the 5-fold cross-validation result was 1.780, and the 10-fold cross-validation result was 1.736. We then created a segmented PLS model based on groups of 50 wavelength points within the 1,600 total points. Both 5-fold and 10-fold cross-validations achieved their lowest error in the second segment, from the 50th to 100th wavelength points, with results of 1.458 for 5-fold and 1.418 for 10-fold cross-validation. We further developed a PLS model combined with a genetic algorithm to select effective wavelength points, and once again, the 10-fold cross-validation outperformed the 5-fold.

For evaluating neural network models, we used 5-fold cross-validation to enhance experimental efficiency. However, we reasonably expect that 10-fold cross-validation could further improve performance in neural networks. Our traditional model results are based on 10-fold cross-validation, while neural network results use 5-fold cross-validation. Even so, the neural network models still significantly outperform the traditional PLS-based models.

### 4.1 Data analysis of variance

First, we applied analysis of variance (ANOVA) to assess the reliability of the fruit dataset we collected. We categorized each fruit sample into three groups based on detected sugar levels: high sugar, medium sugar, and low sugar. This categorization enabled us to statistically evaluate the within-group similarity and between-group dissimilarity.

Categories were defined by specific sugar value thresholds. For instance, in the pear dataset, the mean sugar value of 300 samples is 12.04 with a standard deviation (STD) of 0.95, a maximum value of 15.0, and a minimum value of 8.5. In total, there are 37 unique sugar values across the samples, with each value corresponding to an average of approximately 8.11 pear samples. We defined the low sugar group as samples within the range [8.5, 11.0], the middle sugar group as [11.0, 13.5], and the high sugar group as [13.5, 15.0]. After categorization, the high sugar group had 16 samples, the middle sugar group 232 samples, and the low sugar group 52 samples.

Each fruit sample spectrum comprises 1,600 wavelength points. To simplify the high-dimensional ANOVA process, we conducted ANOVA on each wavelength point individually, then synthesized and averaged the results across all dimensions. Our samples are independent, and each dimension follows a normal distribution. Using pears as an example, we display nine random dimensions of pear spectra in the figures below. For each dimension, we conducted a homogeneity of variance test across the high, middle, and low sugar groups. Dimensions passing the homogeneity of variance test were deemed valid, and only these dimensions were used in ANOVA. We used a 5% significance level.

To ensure the validity of ANOVA, we balanced sample sizes across groups. Since the middle sugar group had more samples than the low, and the low more than the high, we designed an experiment selecting 15 samples per category at random to represent each group. ANOVA was then performed between each pair of categories on all valid dimensions, with results averaged across dimensions. This experiment was repeated multiple times to evaluate variance across the three groups

Second, we analyzed the variation within each sugar value group. Taking the middle sugar group as an example, we divided it into 10 subgroups, each containing approximately 23 samples. We paired these 10 subgroups and performed ANOVA on each pair, averaging the results to statistically assess the within-group similarity of the middle sugar samples.

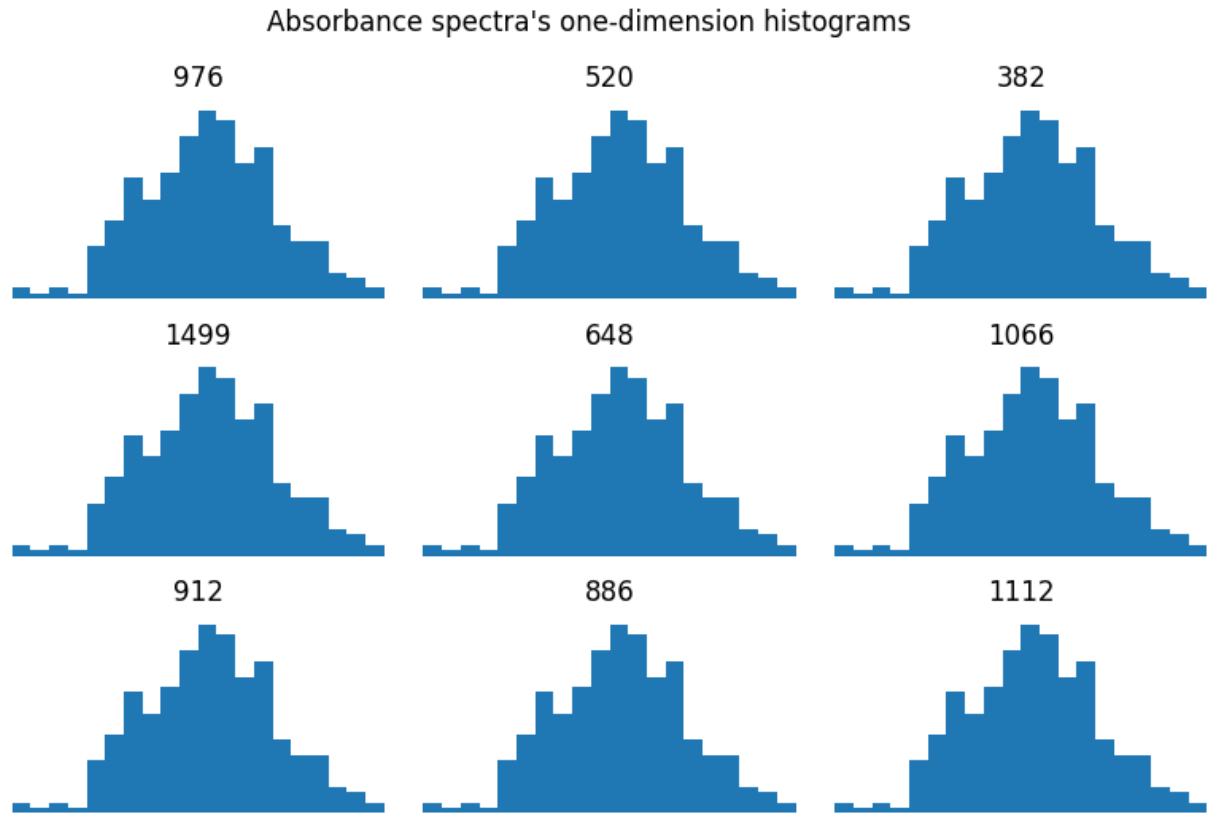

Figure 2. Random dimensions from absorbance spectra.

For pears, our calculations showed that the similarity between the high and middle sugar groups was 18.7%, between the high and low sugar groups was 10.0%, and between the middle and low sugar groups was 38.8%. All these inter-group similarities exceed our standard significance threshold, which could negatively impact the results of subsequent experiments. Thus, we expect that our model will perform better with a dataset exhibiting less similarity between groups. Intra-group similarity results were 70.6% for the high sugar group, 47.3% for the middle sugar group, and 25.8% for the low sugar group.

The results from the Navel orange dataset are similar to those from the pear dataset. The mean sugar value of the 300 Navel orange samples is 14.57, with a standard deviation of 1.64, a maximum of 18.9, and a minimum of 10.2. In total, there are 62 unique sugar values across the samples, with each value corresponding to an average of approximately 4.84 Navel oranges. We categorized the samples as follows: low sugar group [10.2, 13.1], middle sugar group [13.1, 16.0], and high sugar group [16.0, 18.9]. After categorization, the high sugar group contained 69 samples, the middle sugar group 167 samples, and the low sugar group 64 samples.

Calculated similarities between groups were 43.2% for the high and middle sugar groups, 20.5% for the high and low sugar groups, and 23.7% for the middle and low sugar groups. These inter-group similarities also exceed our standard significance threshold. Within each group, the intra-group similarities were 33.4% for the high sugar group, 46.1% for the middle sugar group, and 16.0% for the low sugar group.

Results from both pears and Navel oranges show that the similarity between the middle and low sugar groups is even higher than the intra-group similarity within the low sugar group for both fruits. This suggests that samples from the middle and low sugar groups are easily confusable, with spectra that are challenging to distinguish. However, overall intra-group similarities remain higher than inter-group similarities, indicating that our dataset is still suitable for experimentation.

### 4.2 Research of preprocess strategy

In this paper, we compare various preprocessing methods and a no-preprocessing scenario. Preprocessing methods include multiplicative scatter correction (MSC), Savitzky-Golay smoothing (SG), standard normal variate (SNV), principal component analysis (PCA), first and second-order derivatives, wavelet decomposition (WD), and combinations of these techniques. In the no-preprocessing scenario, our custom neural network model achieved the best performance compared to other models, such as traditional PLS-based and conventional neural network models, with an RMSECV of 0.738. Using this model, we analyzed the effects of different preprocessing methods. Specifically, we applied each preprocessing technique to the spectral data and then fed the processed data into our neural network model for training and evaluation.

First, we applied various preprocessing methods individually. The results indicated that no single preprocessing method outperformed the no-preprocessing scenario. In fact, first-order derivative methods performed noticeably worse than non-derivative methods, and second-order derivative methods performed even worse than first-order derivatives. For example, using first-order derivatives with pears yielded an RMSECV of 0.846, while second-order derivatives resulted in an RMSECV of 1.158. By contrast, all non-derivative preprocessing combinations achieved RMSECVs below 0.750. Consequently, derivative-based preprocessing methods were excluded from further consideration in combinations.

Principal Component Analysis (PCA) also appeared to lead to overfitting, as features extracted by PCA showed significant differences between the training and validation sets, making it challenging to predict corresponding sugar values accurately. Thus, PCA-based preprocessing combinations were not prioritized.

Next, we explored combinations of preprocessing methods, or preprocessing chains, as each method has specific strengths and weaknesses. By combining methods strategically, we aimed to leverage their strengths and compensate for their limitations, potentially capturing intrinsic spectral features. Through repeated experiments, we found that applying Savitzky-Golay (SG) smoothing (using a 5-point least squares analysis), followed by multiplicative scatter correction (MSC), and then standard normal variate (SNV), produced strong results. This combination yielded an RMSECV of 0.722, making it our first-stage preprocessing choice.

Building on these first-stage results, we added wavelet decomposition as a second-stage preprocessing step. Reducing dimensions from 1600 to 400 with wavelet decomposition achieved an RMSECV of 0.716, while reducing to 100 dimensions yielded an RMSECV of 0.724. Although both results were similar, the 400-dimension version preserved more feature details, which proved beneficial for subsequent feature selection using a genetic algorithm. Additionally, wavelet decomposition enhanced model performance and reduced dimensionality, accelerating neural network training and making the model more suitable for real-time detection.

### 4.3 Genetic algorithm model

we add a genetic algorithm (GA) as the third stage of preprocessing. GA can efficiently select key wavelength points from the full spectrum based on Partial Least Squares (PLS), making GA combined with PLS an effective approach to enhance PLS performance.

We began by segmenting the spectrum into equal intervals, testing intervals of 400, 200, 100, and 50. After comparing results, we found that the 50-interval segmentation achieved the best outcome, with an RMSECV of 1.418 (using pear data as an example).

In this stage, we input the 400 features from our second-stage preprocessing into the GA model, using PLS as the objective function to guide feature selection. The GA optimally selected 100 features from the 400, minimizing RMSECV..

Using GA in combination with PLS for optimal feature selection has proven to be a reasonable strategy based on our comparisons. Without the first-stage preprocessing, directly reducing the raw 1600-point spectrum to 100 features yielded an RMSECV of 1.50, an improvement over the non-preprocessed PLS model, which had an RMSECV above 1.70.

When we skipped the first stage but applied wavelet decomposition in the second stage to reduce the spectrum to 100 features, followed by GA to select 20 optimal features, the RMSECV for the second generation was 1.44. Alternatively, using wavelet decomposition to reduce the raw spectrum to 400 features and then applying GA to select 100 features achieved an initial RMSECV of 0.89, which stabilized around 0.82 after 20 generations.

These comparisons demonstrate that wavelet decomposition combined with GA effectively selects features that closely correlate with sugar values. They also highlight that selecting too few features may not adequately represent the spectrum, impacting model performance. The basic structure of our GA model is as follows:

1. Generate the Initial Population: The first generation is considered the mature generation.
2. Evaluate Individuals: Assess each individual's score based on PLS results, where lower scores are preferred.
3. Selection and Pairing: Rank individuals in the mature generation by RMSECV. Select the top 20% based on score and an additional 5% from the remaining 80%. This selected set, representing 25% of the mature generation, forms pairs, with each pair producing 8 offspring.
4. Mutation: Introduce gene mutation in 10% of individuals, with mutation rates following a normal distribution (mean = 0.1, standard deviation = 0.01).
5. Reproduction Cycle: The resulting sub-generation becomes the new mature generation, repeating steps 2 to 5.

Through repeated experiments, we selected a population size of 400 individuals, as a larger population did not significantly improve results or reduce the number of generations needed for RMSECV to stabilize, but did increase reproduction time. Conversely, a smaller population required more generations and increased the risk of underfitting.

In each mature generation, we selected the top 20% of individuals for reproduction to ensure the descent of high-quality genes. Additionally, we selected 5% from the remaining 80% to maintain genetic diversity. Thus, 25% of the mature generation was chosen to reproduce. Each pair produced 8 offspring, keeping the population size at 400. We introduced gene mutation in 10% of the sub-generation, approximately 40 individuals per generation, with mutation rates following a normal distribution (mean = 0.1, standard deviation = 0.01). This means that, on average, $40 \pm 4$ individuals underwent gene mutation each generation.

### 4.4 Neural network model

A new neural network model, named the MLP-CNN model, was proposed based on analysis of spectral features and experimental results. Its structure includes lower layers comprising a Multi-Layer Perceptron (MLP), a middle layer that serves as a connection layer consisting of a 2-dimensional correlation spectrum matrix, and upper layers consisting of Convolutional Neural Network (CNN) layers.

Given the characteristics of chemometrics problems, we observed that spectral features are primarily one-dimensional linear rather than two-dimensional correlations. Therefore, after extensive experimentation, we designed a deep learning regression model with MLP layers. The model begins with six MLP layers, with layer sizes progressively decreasing as follows: 512, 256, 128, 64, 32, and 16 neurons. Each fully connected layer is followed by a ReLU activation layer. The gradual reduction in layer size allows the neural network to learn spectral features effectively.

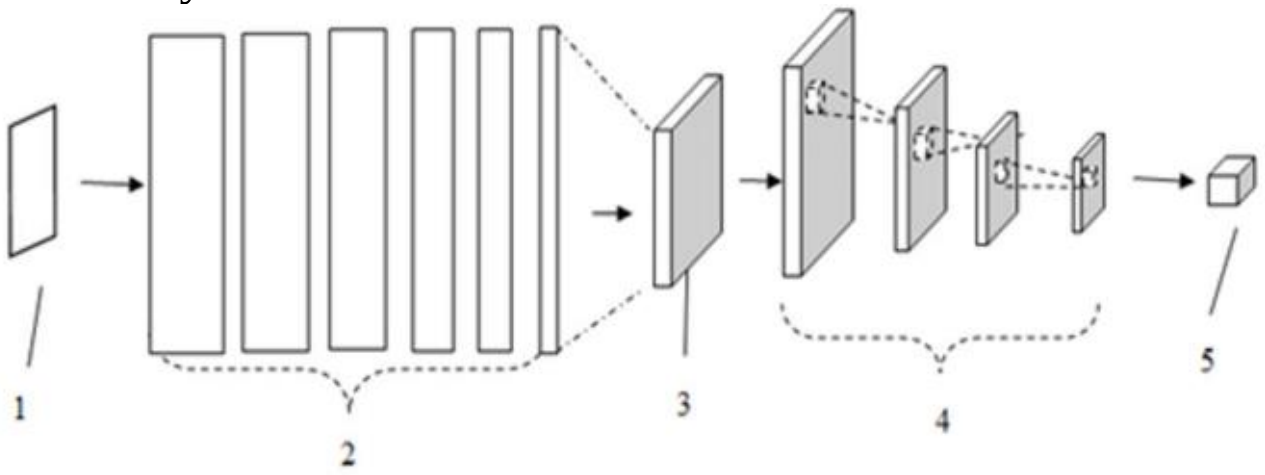

1. Input; 2. MLP; 3. Two-dimensional correlation spectroscopy. 4. CNN; 5. Output.

Figure 3. The structure of MLP-CNN model.

We then use the output features from the sixth fully connected layer as inputs to a 2-dimensional spectral information matrix. These features are highly abstract and resilient to local noise, with each dimension effectively representing the data. The 2-dimensional spectral information matrix is derived from the self-correlation of the 1-dimensional features output by the MLP. The output from this layer is then fed into a 4-layer CNN, with each CNN layer followed by a ReLU activation layer.

The regression model leverages CNN and pooling layers to capture the intrinsic characteristics of spectral features. The network's structure, including local connectivity and weight sharing, reduces the number of independent parameters and enhances model generalization. The filter sizes of the CNN layers progress as 64, 64, 128, and 128, with convolutional kernel sizes of 3x1, 1x3, 3x1, and 1x3, respectively. This combination of small kernel sizes achieves similar performance to more complex kernels, increases nonlinearity, reduces kernel parameters, and provides implicit regularization. The output from the fourth CNN layer feeds into the model's final layer, a special CNN layer. Inspired by global average pooling, we designed this layer with a kernel size matching the input size of this layer, producing a single output representing the predicted sugar value.

The 2-dimensional spectral information matrix represents an adjustment in the neural network structure that preserves the information from MLP output features while extending 1-dimensional linearity into 2-dimensional correlations. Since this matrix is derived from the self-correlation of 1-dimensional features, it is a real symmetric matrix. This property allows for diagonalization, with different eigenvectors corresponding to distinct eigenvalues that are orthogonal. This orthogonality makes the matrix convenient for further processing and analysis.

In this paper, we consider spectral data to have strong 1-dimensional relevance. We therefore designed an innovative model structure with an MLP at the beginning and a CNN at the end, using self-correlation to convert data into a 2-dimensional matrix format. CNNs are well-documented to effectively learn spatial correlations in 2-dimensional data, as well as local features through convolution and pooling layers.

If an alternative strategy were used, such as directly arranging 1-dimensional features into a matrix by wrapping each filled row to the next, important correlations—such as those between the end of one row and the beginning of the next—would be lost. For example, in a column, adjacent elements in the original 1-dimensional features would be separated by an entire row, weakening the learned representation of original spectral features. In one-dimensional spectra, close proximity often indicates high relevance, a characteristic that CNNs cannot capture well without self-correlation.

By applying a self-correlation operation to the 1-dimensional data, we create a 2-dimensional spectral information matrix that naturally extends linearity into 2-dimensional correlations, thus expanding the feature space. CNNs can then effectively learn these 2-dimensional correlations, accurately representing the original spectral data.

Given our sample size of only 300, we set the training batch size to the entire training set to ensure thorough learning of features related to different sugar values. For example, with 240 samples as the training set and 60 as the validation set, each training batch contains 240 samples. The training and validation sets are distinct from one another.

### 4.5 Evaluation standards

In this paper, we propose a new evaluation standard for sugar value detection: the effectiveness of detection is assessed by the ratio of the prediction results' root mean square error (RMSECV) to the dataset's standard deviation. Unlike traditional evaluation standards that rely solely on the absolute RMSECV value, this ratio-based approach provides a more balanced assessment. We refer to this ratio as *Closeness*.

$$\mathrm{RMSEP} = \sqrt{\sum_i (\mathrm{Y}_i^{predict} - \mathrm{Y}_i^{true})^2} \quad (1)$$

$$STD = \sqrt{\sum_i (\mathrm{Y}_i^{true} - Y_{mean})^2} \quad (2)$$

$$Closeness = \frac{RMSECV}{STD} \quad (3)$$

Traditional evaluation standards that rely solely on the absolute RMSECV value may not provide a reliable measure of model generalization, as different fruit datasets often have varying sugar value ranges and variances. Instead, using the standard deviation (STD) of the dataset's sugar values as a scaling factor allows for a more meaningful comparison across datasets.

For example, using our Navel orange and pear datasets with the neural network model we designed, the RMSECV for Navel oranges is 1.184, and for pears, it is 0.710. These absolute values alone do not reveal the model's relative performance. However, when we factor in the dataset STD, the results become clearer: the STD for Navel oranges is 1.642, and for pears, it is 0.955. By calculating the ratio of RMSECV to dataset STD, we find a *Closeness* value of 72.1% for Navel oranges and 74.3% for pears, indicating that our model generalizes well between these two fruit types.

For PLS-based models, the best results were achieved by combining PLS with GA for optimal feature selection. In this case, the Closeness value for the Navel orange dataset is 86.5%, and for the pear dataset, it is 90.4%. These similar ratios further support the generalizability of the model.

Traditional evaluation standards often include the coefficient of determination ($R^2$). Using the pear dataset as an example, our designed neural network model achieved the lowest RMSECV. However, the average $R^2$ across the 5-fold cross-validation results was 0.314, indicating that only 31.4% of the variance in sugar value can be explained by the linear relationship between fruit spectra and sugar content. The remaining 68.6% is influenced by other, less clear factors.

We believe that with more precise experimental apparatus, a larger fruit sample size, and enhanced computational power, it would be possible to achieve better results without altering the model strategy or neural network architecture we proposed.

## 5. Results

### 5.1 Comparison of different preprocess methods

First, we applied each preprocessing method individually. Next, we tested combinations of these individual preprocessing methods. Finally, we compared the performance of all preprocessing approaches.

**Table 1. Spectra regression strategies based on MLP-CNN (Pears).**

| Spectra regression strategies | RMSECV |
|---|---|
| Non > MLP-CNN | 0.738 |
| SG > MLP-CNN | 0.748 |
| MSC > MLP-CNN | 0.722 |
| SNVC > MLP-CNN | 0.720 |
| WD (400) > MLP-CNN | 0.716 |
| WD (100) > MLP-CNN | 0.724 |
| SG > MSC > SNV > MLP-CNN | 0.722 |
| SG > MSC > SNV > WD (400) > MLP-CNN | 0.746 |
| SG > MSC > SNV > WD (100) > MLP-CNN | 0.720 |
| SG > MSC > SNV > WD (400) > GA (100) > MLP-CNN | 0.710 |
| SG > MSC > SNV > WD (100) > GA (25) > MLP-CNN | 0.722 |

As shown in the table, the optimal preprocessing strategy for our neural network model is: SG > MSC > SNV > WD (400) > GA (100) > MLP-CNN. Using pears as an example, our MLP-CNN model performs well even without preprocessing. However, the designed preprocessing strategy further enhances performance.

All tested combinations achieved an RMSECV in the 0.72 range, except for SG and SG > MSC > SNV > WD (400) > MLP-CNN. Thus, preprocessing combinations improved the MLP-CNN model's performance by approximately 0.2 RMSECV.

### 5.2 Comparison of neural network and PLS based model

The performance of the proposed neural network model, traditional PLS-based models, and the PLS combined with GA model were compared.

Experimental results indicate that the neural network models significantly outperform the PLS-based models. Using pears as an example, the original PLS model achieved an RMSECV of 1.736. The optimal preprocessing strategy we identified is SG > MSC > SNV > WD (400) > GA (100). With this strategy, the PLS model combined with GA achieved an RMSECV of 0.82 after the 20th generation.

When we fed the 100 features output by this preprocessing combination into our neural network model, the RMSECV was reduced to 0.710. Eliminating the last stage of preprocessing (GA) slightly increased the RMSECV to 0.746, indicating that wavelet decomposition (WD) is a crucial step. The final feature set produced by GA is highly representative, reducing the number of features from 400 to 100 and delivering better performance than directly reducing to 100 features with WD alone (RMSECV = 0.724).

**Table2.Comparison of PLS based or MLP-CNN based strategies.**

| Spectra regression strategies | RMSECV |
|---|---|
| Non > PLS | 1.736 |
| Non > Equal interval segment PLS (50) | 1.418 |
| SG > MSC > SNV > WD (400) > GA (100) > PLS | 0.827 |
| SG > MSC > SNV > WD (100) > GA (20) > PLS | 1.444 |
| SG > MSC > SNV > WD (400) > GA (100) > MLP-CNN | 0.710 |

### 5.3 Comparison between different neural network models

In this paper, we compared our proposed MLP-CNN model with models using only the MLP or CNN components, as well as the traditional CNN-MLP structure. The MLP-CNN model consistently achieved approximately a 75% ratio of RMSECV to dataset STD across both fruit types (Navel orange and pear), regardless of the preprocessing method or combination used. Different preprocessing strategies affected RMSECV by only ±5%.

The RMSECV of the MLP-CNN model reached 0.710. When using only the MLP component of the model, the RMSECV also reached 0.710. However, using only the CNN component without applying self-correlation resulted in an RMSECV of 0.748. In this case, the input feature dimension was 100, and self-correlation would produce a 100x100 matrix per fruit sample, which is too large to train efficiently. Training this matrix format for 5,000 epochs on a GTX1060 GPU took approximately 45 minutes, making it less practical. Without self-correlation, simply reshaping the 100 features into a 10x10 matrix reduced the training time for 5,000 epochs to 70 seconds.

Among the models, the MLP-only model trained the fastest, completing 5,000 epochs in about 30 seconds. The proposed MLP-CNN model required approximately 96 seconds for 5,000 epochs, which remains efficient and manageable.

An interesting phenomenon emerged when training our proposed neural network model on PCA-preprocessed features: the RMSECV of the training set dropped below 0.1 within 1,000 epochs, indicating rapid overfitting. In contrast, using only the MLP or CNN components did not lead to overfitting as quickly, even with the small dataset, and showed a lower overall degree of overfitting. This suggests that our model learns low-dimensional features faster and more accurately.

However, with only 300 samples per fruit, the dataset lacks sufficient diversity in spectral patterns. In real scenarios, fruits with close or identical sugar values can exhibit a wide variety of spectral shapes. As a result, PCA preprocessing yields principal components that differ significantly across samples, making them poor representations of each other. This leads to severe overfitting on the training set and a high RMSECV on the validation set—around 130% when normalized by dataset STD—similar to results from equally segmented PLS models.

We believe that expanding the dataset to 3,000 or even 30,000 samples would significantly improve model performance without requiring any changes to the model architecture

**Table 3. Comparison of MLP-CNN-based designs with other neural networks.**

| Spectra regression strategies | RMSECV |
|---|---|
| Non > MLP | 0.780 |
| Non > CNN | 0.730 |
| Non > CNN-MLP | 0.743 |
| Non > MLP-CNN | 0.748 |
| SG > MSC > SNV > WD (400) > GA (100) > MLP | 0.710 |
| SG > MSC > SNV > WD (400) > GA (100) > CNN | 0.748 |
| SG > MSC > SNV > WD (400) > GA (100) > CNN-MLP | 0.735 |
| SG > MSC > SNV > WD (400) > GA (100) > MLP-CNN | 0.710 |

## 6. Conclusion

Experimental results indicate that, for both Navel orange and pear datasets, applying wavelet decomposition to reduce spectral data dimensions followed by GA for optimal feature selection proves effective. The preprocessing strategy SG > MSC > SNV > WD (400) > GA (100) > MLP-CNN is the most efficient approach for sugar value detection in our dataset. Using the Closeness metric we developed, this strategy achieved a score of 75.0% ± 5.0%. In certain specific cases, targeting particular training and test sets, the Closeness value of the MLP-CNN model reached as low as 15.0%, suggesting promising directions for further research.

Overall, the neural network models performed significantly better than the PLS-based models, with the PLS combined GA model being the strongest among the PLS models. Additionally, MLP models outperformed CNN models, likely due to the predominantly 1-dimensional linear nature of fruit spectra, as opposed to 2-dimensional correlations.

Most importantly, we demonstrated that the performance of our proposed MLP-CNN model matches that of the MLP-only model and exceeds that of the CNN-only model. This suggests that our MLP-CNN model has strong potential for applications in fruit soluble solid content detection, and further research on this model is worthwhile.